\pdfoutput=1
\documentclass[11pt]{article}

\usepackage{acl}

\usepackage{times}
\usepackage{latexsym}

\usepackage[T1]{fontenc}
\usepackage[utf8]{inputenc}
\usepackage{microtype}
\usepackage{xcolor}
\usepackage{geometry}
\usepackage{adjustbox}
\usepackage{float}
\usepackage{placeins}
\usepackage{acronym}
\usepackage{csquotes}
\usepackage{graphicx}
\usepackage{caption}
\usepackage{subcaption}
\usepackage{todonotes}
\usepackage{threeparttable}
\newcommand{\commentout}[1]{}

\newcommand{\anon}[1]{#1} % unmasked
\newcommand{\sarg}[1]{\texttt{#1}}

\newacro{BNC}{British National Corpus}
\newacro{SRL}{semantic role labeling}
\newacro{NLP}{natural language processing}
\newacro{RoFA-MT}{role-filler averaging model}
\newacro{ResRoFA-MT}{residual role-filler averaging model}
\newacro{NNRF}{non-incremental role-filler model}
\newacro{NNRF-MT}{non-incremental role-filler multitask model}
\newacro{PReLU}{parametric rectified linear unit}
\newacro{POS}{parts-of-speech}
\newacro{OOV}{out-of-vocabulary}
\newacro{WSJ}{Wall Street Journal}
\title{Where's the Learning in  Representation Learning for Compositional Semantics and the Case of Thematic Fit}

\author{Mughilan Muthupari$^*$ \\
  Columbia University \\
  NY, USA \\
  \texttt{mm5569@columbia.edu} \\
  \\
    \textbf{Asad Sayeed} \\
  Dept. of Philosophy, Linguistics,\\
  and Theory of Science \\
  University of Gothenburg,
  Sweden \\
  \texttt{asad.sayeed@gu.se} \\
  \And
  Samrat Halder$^*$ \\
  Columbia University \\
  NY, USA \\
  \texttt{sh3970@columbia.edu} \\
  \\
  \textbf{Yuval Marton} \\
  University of Washington \\
  WA, USA\\
  \texttt{ymarton@uw.edu} \\  
  }

\begin{document}
\maketitle
% anonymized. uncomment for camera-ready
\def\thefootnote{*}\footnotetext{These authors contributed equally to this work}

%! Author = mughi
%! Date = 2/25/2022

\begin{abstract}
    Observing that for certain NLP tasks, such as semantic role prediction or thematic fit estimation,
    random embeddings perform as well as pretrained embeddings,
    we explore what settings allow for this and examine where most of the learning is encoded: the word
    embeddings, the semantic role embeddings, or ``the network''.
    We find nuanced answers, depending on the  task and its relation to the training objective.
    We examine these representation learning aspects in multi-task learning, where role prediction and role-filling are  supervised tasks, while several thematic fit 
    tasks are outside the models' direct supervision.
    We observe a non-monotonous relation between some tasks' quality score  
    and the training data size.
    In order to better understand this observation, we analyze these results using easier, per-verb versions of these tasks.
\end{abstract}

%! Section 1 - Introduction

\section{Introduction}
\label{sec:introduction}  % Use \autoref to reference the section everywhere else.

%------------------------------------------------------------------------------------------------------------------

We examine to what extent models trained on a simplified \ac{SRL} task can estimate thematic fit
(aka semantic fit), as the training set size grows -- and where most of the learning is stored:
in the word embeddings, the thematic role embeddings, or elsewhere in the neural net.

A major goal of \ac{NLP} is to understand the semantics of language. One traditional \ac{NLP} task
around this is 
%semantic role labeling (SRL)
\ac{SRL}, which labels word spans in a sentence with thematic roles. Consider the sentence
\enquote{I cut the cake with a knife}. We can interpret \enquote*{cut} as the action, \enquote*{I} as the \sarg{Agent} 
(the performer of the action), \enquote*{cake} as the \sarg{Theme} of the action (the thing that underwent the action), and \enquote*{knife} as the \sarg{Instrument} of the action. 
%We will use the above interpretation of our example sentence. This sentence represents an event that took place, and the sentence’s event representation consists of the action, the objects related to the action, and their corresponding roles. 
These words, labeled with roles such as \sarg{Agent}, \sarg{Theme}, and \sarg{Instrument}, would be our representation 
of the event that the sentence conveys. 
%(We will use a simplified role set instead: \sarg{Arg0}, \sarg{Arg1}, etc. See Section~\ref{sec:methodology}).
Other sentences with similar meanings, e.g., 
\enquote{the cake was cut with the knife by me}, should have the same (or very similar) event representations.
In this work, we focus on model training with a simplified version of \ac{SRL}: each event is represented only by the lemmatized syntactic head 
of each event argument (including the predicate), and the semantic roles are the simplified PropBank 
roles (\sarg{Arg0, Arg1}, etc.). 
The reason for this is the current limitations of available evaluation sets for thematic fit: 
they are all comprised of lemmatized syntactic argument heads as well.

\textbf{Thematic fit} is related to \ac{SRL}.
%, but separate.
This task aims to identify how well a given word or concept fits into a role of an event. 
%Going back to 
In
our example sentence, consider these potential replacements for \enquote*{knife}: scissors, fork, and brick. 
As humans, we understand that while \enquote*{knife} is the most typical object for this situation, both \enquote*{scissors} and 
\enquote*{fork} could also fit, even if not as naturally. 
This is because we have the general intuition that all three objects are plausible instruments for cutting. 
More so, we know that \enquote*{brick} is unlikely to fit given the context of cutting a cake. 
%This is an example of what thematic fit attempts to solve. 
Since thematic fit datasets are scarce, one challenge in computational linguistics (and computational psycholinguistics)
revolves around how machine learning models can learn thematic fit indirectly -- perhaps from \ac{SRL} training.
% \todo{mention theoretical implications for human learning of sem fit?}
To the best of our knowledge, the state-of-art in this line of work is the \ac{ResRoFA-MT} proposed 
by \citet{hong-2018-learn-event-rep}, with an adjusted embeddings representation and training data annotation in 
\citet{marton-sayeed-lrec2022-RW-eng-v2}.

It has been repeatedly observed that in some settings, random word embeddings perform as well as pretrained ones, 
or very nearly, including in our baselines \cite{tilk2016event,hong-2018-learn-event-rep,marton-sayeed-lrec2022-RW-eng-v2}. In this paper, we design experiments to answer the following questions: \textbf{Q1.} 
Why is this so
in our compositional semantics and psycholinguistic tasks? 
\textbf{Q2.} For such semantic tasks and architecture, where is the learning encoded? Is it in the word embeddings, 
role embeddings, or elsewhere in the neural network?
\textbf{Q3.} Training set size effect: is more data better for this indirect setting and tasks? 

% In this paper, we examine training set size effects on thematic fit tasks -- for which the models were not directly 
% optimized -- even after reaching a plateau on the simplified \ac{SRL} task and its complementary task 
% (predicting the head word given the role). 
% \textbf{1.}~We find surprising training set size interactions with specific evaluation sets and design a modified evaluation metric in order to better understand these interactions.
%
In this work,
\textbf{1.}~We compare updating the word embeddings during training to freezing them. 
\textbf{2.}~We modify the \ac{ResRoFA-MT} model architecture in various ways to understand what contributes the most to the learning: 
 the pretrained (or random) word embeddings, the thematic role embeddings, or the rest of the network.
\textbf{3.}~In order to be able to train on larger data, we optimized the code of
\citet{hong-2018-learn-event-rep} and \citet{marton-sayeed-lrec2022-RW-eng-v2}.
We release our optimized codebase\footnote{\anon{\url{https://github.com/MughilM/RW-Eng-v3-src/tree/arxiv_release}}}, which trains 6 times faster
and includes ablation architectures and a correction to the training data preparation step.

%! Section 2 - Related Work

\section{Related Work}
\label{sec:related-work}
%\todo{Asad: pls add/ modify}

%-----------------------------------------------------------------------------------------------------

In event representation models, the main goal is to predict the appropriate word in a sentence given both the role of that
 word and the surrounding context in the form of word-role pairs. 
One of the best early neural models was the \ac{NNRF},  by \citet{tilk2016event}.
This model was based on selectional preferences, or a probability distribution over the candidate words. 
However, one drawback of this model is that representations of two similarly-worded sentences differing hugely
in meaning would closely resemble each other, e.g., \enquote{kid watches TV} and \enquote{TV watches kid}.
Another drawback is that 
 the embeddings of the word-role pairs are summed together to represent the sentence, and so the resulting event representation vector does not weight the input vectors 
differently based on their importance and is not normalized for varying numbers of roles in a sample.

\citet{hong-2018-learn-event-rep} extend this model in three ways: 
First, in addition to the word prediction task of \ac{NNRF}, the task of role prediction given the corresponding word is added, 
and the two tasks are trained simultaneously (multi-task learning). 
This model is known as the \ac{NNRF-MT}.
Second, they apply the \ac{PReLU} non-linear function to each word-role embedding, which acts 
as weights on the composition of embeddings, and subsequently average the embeddings, which normalizes for variable 
length inputs. 
This model is called the \ac{RoFA-MT}. 
Third, in an effort to tackle the vanishing gradient problem, residual connections between the \ac{PReLU} output and the 
averaging input were added together. 
This third iteration is known as the \ac{ResRoFA-MT} model. 
They showed that it performs the best on our thematic fit tasks, and so we use it as our baseline.

Our work differs from \citet{hong-2018-learn-event-rep} and  \citet{marton-sayeed-lrec2022-RW-eng-v2}
in that while they focused more on state-of-the-art performance  through new modeling and 
annotation methods, we aim to understand what controls the learning in such networks.
Also, 
although \citeauthor{hong-2018-learn-event-rep} confirm in private communication that they found pre-trained and random embeddings performance similar in preliminary studies,
none of the surveyed previous work published experiments with pre-trained embeddings. We are the first to do so (using GloVe) and compare that to using random embeddings.

Previous work suggests a difference between "count" and "predict" models, where "count" models represent lexical semantics in terms of raw or adjusted unsupervised frequencies of correlations between words \citep[such as Local Mutual Information;][]{baroni2010distributional} and syntactic or semantic phenomena;
"predict" models involve supervised training to achieve their representations, e.g., neural models.  \citet{baroni2014don} do a systematic exploration of tasks vs. state-of-the-art count and predict models and find that predict models are overall superior; for thematic fit, predict models are the same or better than count models on the best unsupervised setup for the task, although they are easily beaten by third-party baselines based on supervised learning over count models. More recently, \citet{lenci2022comparativeDSM} demonstrate that predict-models are not reliably superior to count-models, but depend on the task and the way the models are trained. They also show that even recent contextual models (e.g., BERT) are not necessarily better for out-of-context tasks than well-tuned static representations, predict or otherwise.
See Appendix A for details on why we do not use BERT in this work.

%! Section 3 - Datasets

\section{Datasets}
\label{sec:datasets}

We use the Rollenwechsel-English, Version 2 (RW-Eng v2) corpus \cite{marton-sayeed-lrec2022-RW-eng-v2} as the training set for all our experiments.
This corpus is sentence-segmented, annotated with morphological analyses, syntactic parses, and syntax-independent PropBank-based \acf{SRL}.
The syntactic head word of each semantic argument is determined by using several heuristics to match the parses 
to the semantic argument spans. 
Note that a sentence may have multiple predicates (typically verbs) and therefore multiple semantic frames 
(sometimes called \enquote{events}), each with its own semantic arguments, whose span may overlap the argument span of 
other frames in the sentence.

The first version of this corpus contained NLTK lemmas, MaltParser parses, \ac{POS}
tags, and SENNA \ac{SRL} tags~\cite{bird2006nltk,nivre2006maltparser,senna}.
The second version added layers from more modern taggers: Morfette lemmas, spaCy syntactic parses and \ac{POS} tags, 
and LSGN \ac{SRL} tags~\cite{chrupala-2011-efficient-morfette,spacy-honnibal-johnson:2015:EMNLP,lsgn-he-2016-2018}.
In our experiments here we use the lemmas of the semantic arguments' head words in v2.

The sentences themselves are taken from both the ukWaC \cite{ukWac} and the \ac{BNC}.
This corpus contains ~78M sentences across ~2.3M documents. 
This includes ~210M verbal predicates with ~700M associated role-fillers.
We use the same training, validation, and test split as \citet{hong-2018-learn-event-rep}. That is, we have 99.2\% (~201.5M samples) in the full training set, 0.4\% in validation, and 0.4\% in testing. 
We run our training experiments on different subsets of the training data, ranging from 0.1\% up to the full dataset. We cap our vocabulary size at the 50,000 most common words in that specific subset.

We used the following psycholinguistic test sets:

\paragraph{Padó}
\cite{pado2006combining}
414 verb-argument pairs and the associated judgement scores. These were constructed from 18 verbs that are present in both FrameNet and PropBank.  For each verb, the three most frequent subjects and objects from each of the underlying corpora were selected.  This process yielded six arguments per verb per corpus, with some overlap between corpora.  For each verb-argument pair, a judgement was collected online with an average of 21 ratings per item for the argument in subject and object role. The rating was collected on a Likert scale of 1-7 with the question "How common is it for [subject] to [verb]?" or "How common is it for [object] to be [verbed]?"

\paragraph{McRae} 
\cite{mcrae1998modeling}
1444 pairs of verb-argument pairs in a similar format to Padó.  These were created using a similar rating question as the Padó dataset, but is  a compilation of ratings collected over several studies with considerable overlap and heterogeneous selection criteria.

\paragraph{Ferretti-Instruments and Ferretti-Locations} 
\cite{ferretti2001}
274 predicate-location pairs and 248 predicate-instrument pairs. Based on the McRae dataset (Psychological norms).

%\paragraph{Greenberg (GDS)} 
\paragraph{GDS} 
\cite{greenberg2015verb}
720 predicate-object pairs and their ratings.  Only objects (no subjects),
matched for high and low polysemy and frequency, well fitting vs.~poorly fitting.
Greenberg and McRae overlap by about a third, but the human scores are obtained from new surveys.

\paragraph{Bicknell}
\cite{bicknell2010effects}
64 cases.
Congruent vs incongruent \sarg{Patient} in an \sarg{Agent}-Verb-\sarg{Patient} paradigm. Hand crafted, not corpus-based, designed for event-related potentials-based neurolinguistic experiments.

%! Section 4 - Modeling and Methodology

\section{Modeling and Methodology}
\label{sec:methodology}

%-----------------------------------------------------------------------------------

In this setup, an input event is represented as role-word pairs, 
where the role is one of the following PropBank \cite{propbank-PalmerGildeaKingsbury:2005} roles:
\sarg{Arg0, Arg1, ArgM-Mnr, ArgM-Loc, ArgM-Tmp}, and the predicate.
The word is the argument's syntactic head's lemma. 
Both the role and the head word are taken from RW-Eng v2.\footnote{Note the input is \textit{not} a full sentence, precluding the use of contextual models such as BERT. See Appendices for details.}
\commentout{
    Each word embedding is multiplied with its corresponding role embedding, and these word-role pairs are combined to a 
    context layer, and a residual connection. 
    In a multitask learning setting for predicting either the next argument's syntactic head word given the role, 
    or the role given the word -- following the \ac{ResRoFA-MT} architecture in \citet{hong-2018-learn-event-rep}, 
    which is also used in \citet{marton-sayeed-lrec2022-RW-eng-v2}.
}

\begin{figure*}[ht!]
    \begin{subfigure}{0.475\textwidth}
        \centering
        \includegraphics[width=\textwidth]{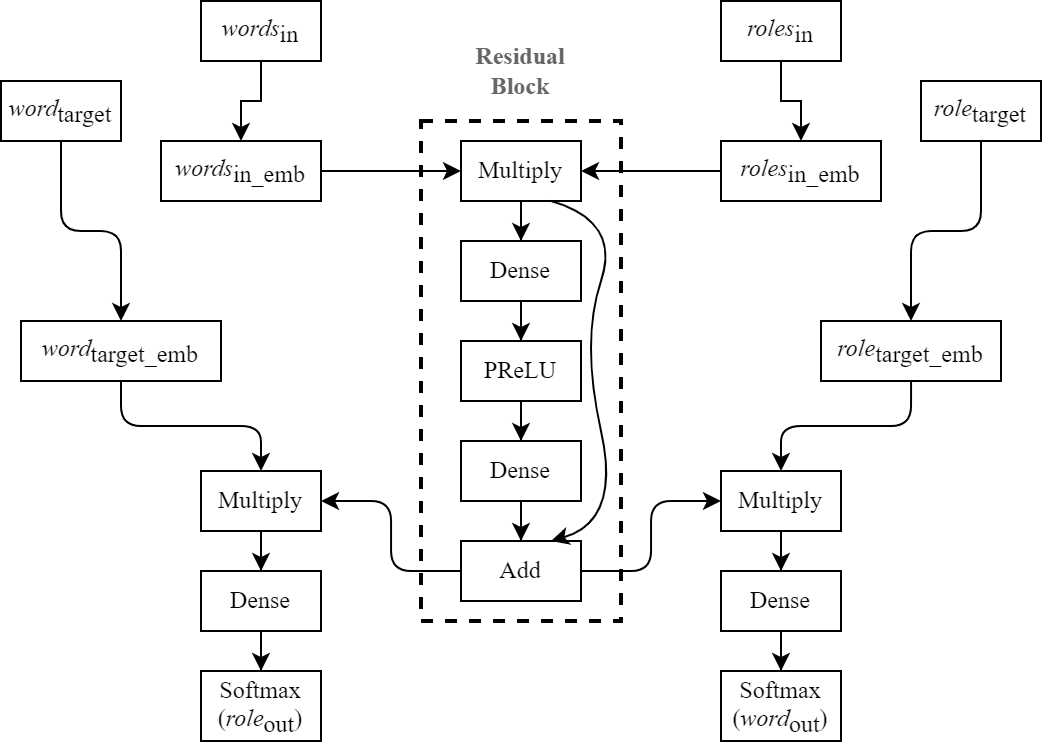}
        \vspace{.01cm}
        \caption{Original \ac{ResRoFA-MT} architecture}
        \label{fig:v4}
    \end{subfigure}
    \hfill
    \begin{subfigure}{0.475\textwidth}
        \centering
        \includegraphics[width=\textwidth]{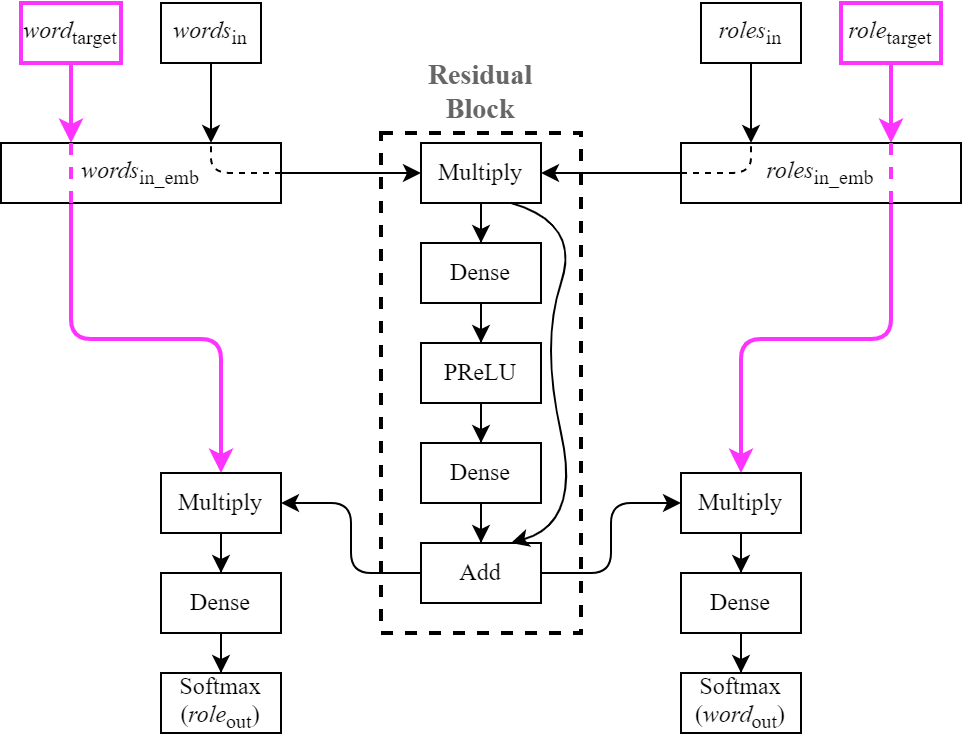}
        \caption{Shared Embedding Layer}
        %{MTRFv5Res with a reduced parameter space.}
        \label{fig:v5}
    \end{subfigure}
    \vspace{.1in}
    
    \begin{subfigure}{0.475\textwidth}
        \centering
        \includegraphics[width=\textwidth]{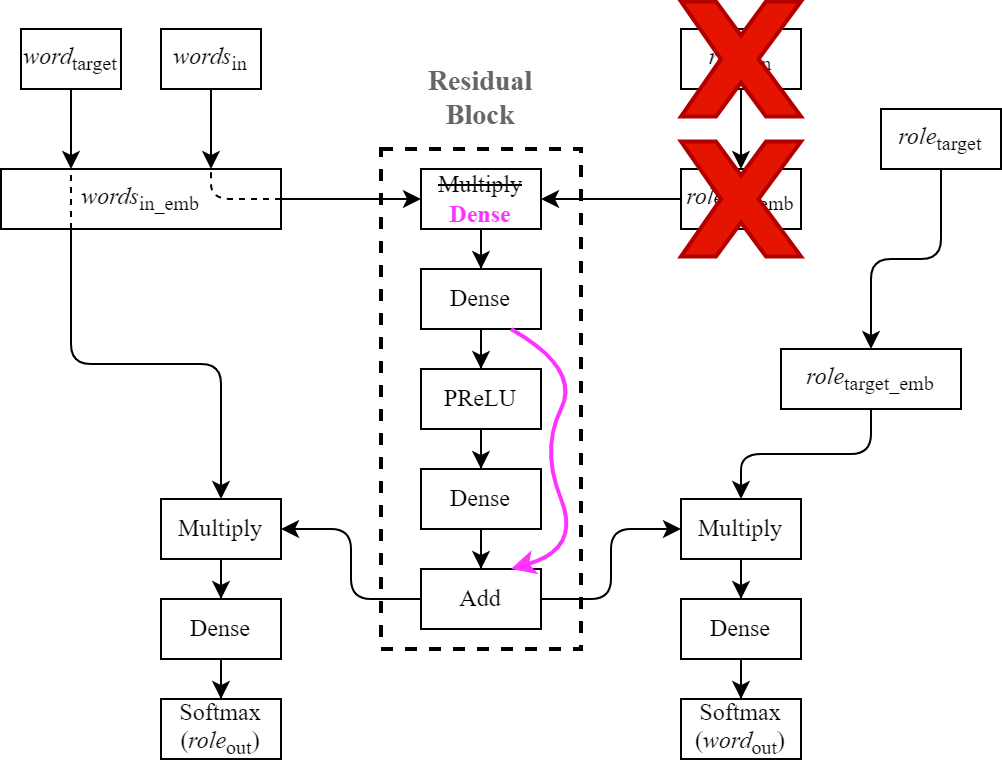}
        \caption{\textbf{NIR} network}
        \label{fig:v6_n1}
    \end{subfigure}
    \hfill
    \begin{subfigure}{0.475\textwidth}
        \centering
        \includegraphics[width=\textwidth]{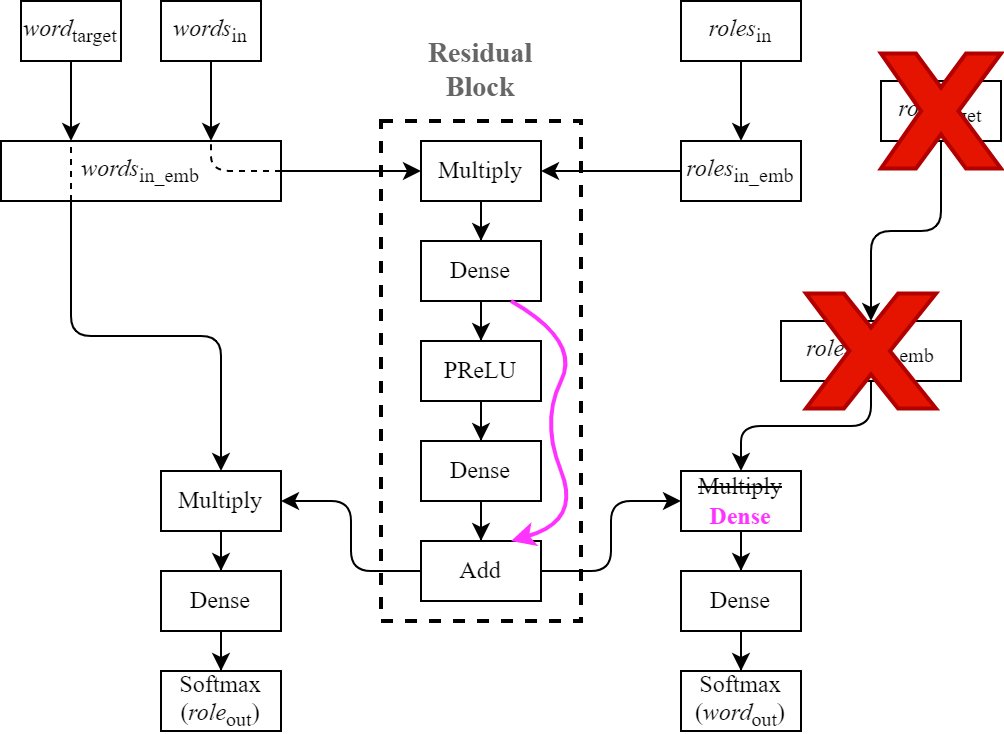}
        \vspace{.01cm}
        \caption{\textbf{NTR} network}
        \label{fig:v6_n2}
    \end{subfigure}
    \vspace{.1in}
    
    \begin{subfigure}{0.475\textwidth}
        \centering
        \includegraphics[width=\textwidth]{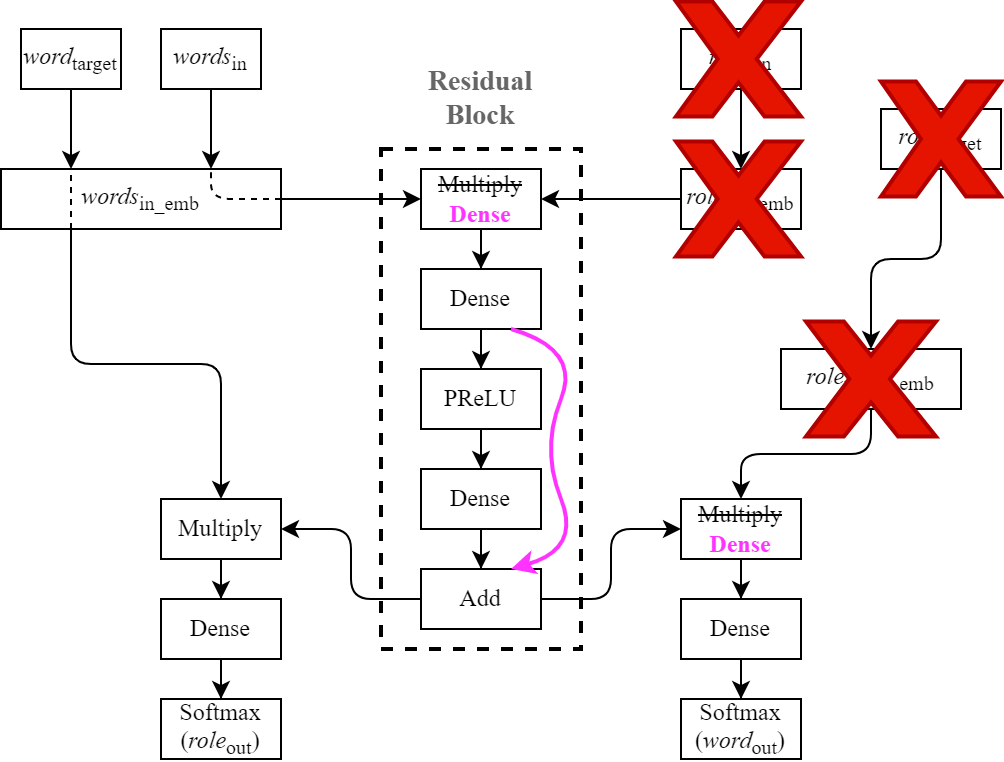}
        \caption{\textbf{NR} network}
        \label{fig:v6_n3}
    \end{subfigure}
    \hfill
    \begin{subfigure}{0.475\textwidth}
        \centering
        \includegraphics[width=\textwidth]{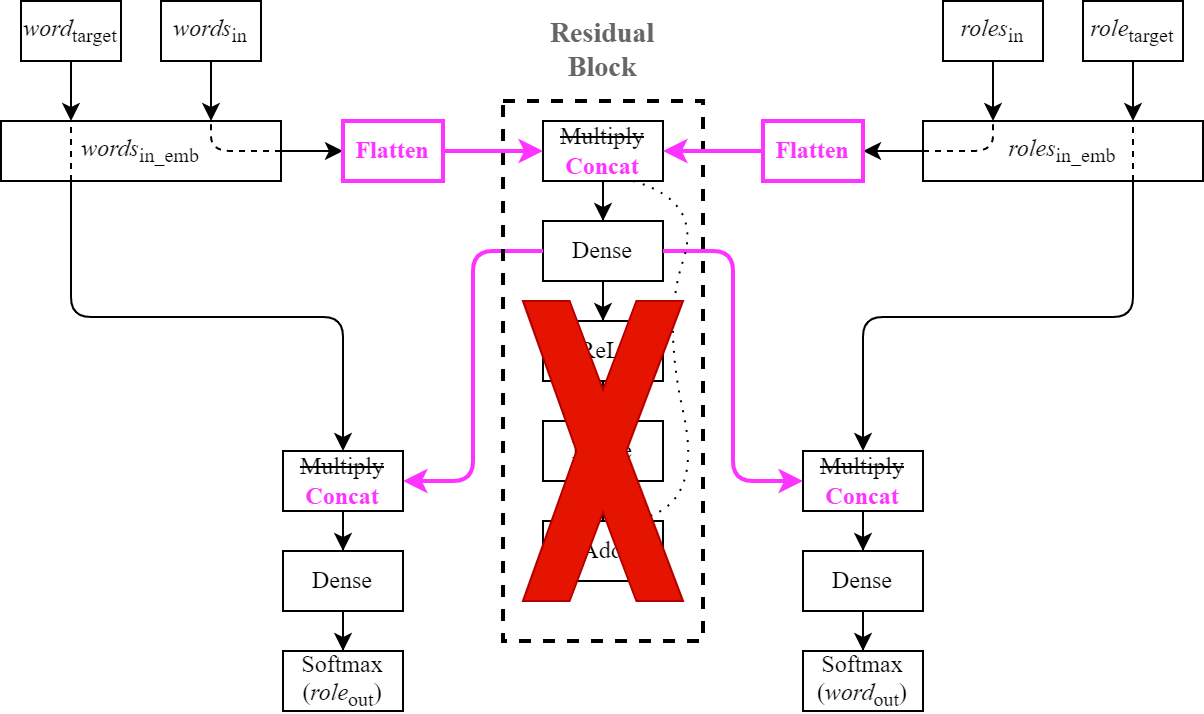}
        \vspace{.5cm}
        \caption{Simple network}
        \label{fig:v9}
    \end{subfigure}
    \caption{The model architectures for our experiments}
    \label{fig:allnets}
\end{figure*}

% \subsection{Objective and Evaluation}
    We train a feed-forward network in a multi-task learning setting to optimize word and role prediction accuracy. For target word prediction we give the prediction layer the target role and a context vector formed as a multiplication of the input word--role pairs. Similarly, for target role prediction we feed the same context vector along with the target word, following the \ac{ResRoFA-MT} architecture \cite{hong-2018-learn-event-rep}
    %,marton-sayeed-2021-RW-eng-v2} 
    (Figure \ref{fig:v4}).
    Since the network initialization is random, we perform~5 runs of each experiment and report the mean with a 95\% confidence interval.
    Following  \citet{hong-2018-learn-event-rep,marton-sayeed-lrec2022-RW-eng-v2}, we test each model on the psycholinguistic datasets (Section~\ref{sec:datasets}),  
    for which the models were not directly optimized.  
    The idea behind using the latter test battery is that the model, even though trained on (simplified) \ac{SRL} and 
    word prediction (aka role-filling) tasks,
    is expected to be able to make indirect generalizations about predicate--argument 
    fit level from the training data and the related objectives.
    These psycholinguistic tasks  
    %Padó, McRae, GDS and Ferretti 
    are evaluated with Spearman's rank correlation %\todo{cite \cite{spearman}} 
    between the sorted human scores and the sorted model scores, 
    except for Bicknell, for which we take accuracy of predicting which argument in each \sarg{Patient} role-filler pair is (more) congruent 
    \cite{LenciECU}.

All prior work with the ResRoFA-MT model uses two random word embedding sets (one for input words and one for the target word) and similarly two role embedding sets. See Figure~\ref{fig:v4}.

Our implementation differs in these key aspects:
\begin{itemize}
\setlength\itemsep{-0.3em}
    \item \textbf{Modified model architecture} - Using a single word embeddings set, shared between the target and input words, and similarly a single role embeddings set (Section~\ref{sec:shared-emb}, Figure~\ref{fig:v5}). 
        In our experiments, we find the non-shared, redundant embedding layers do not affect the performance while adding (vocab size 50,000 $\times$ word embedding size 300) 15,000,000 learnable parameters in the model. 

    \item \textbf{Changes in Batching} - With previous implementations, one \textit{epoch} only resulted in about a third of the data being traversed. 
        The next epoch would start on the second third and so on. 
        Now, we set the data preprocessing so that 
        one \textit{epoch} is one pass through all the training data. 
        Additionally, the data is preprocessed during the training of each batch, so no time is lost during training in waiting for the next batch of data to be preprocessed.
    
    \item \textbf{Missing and unknown words handling} - Following \citet{marton-sayeed-lrec2022-RW-eng-v2} but 
        unlike \citet{hong-2018-learn-event-rep}, we represent \ac{OOV} words separately 
        from missing words (empty slots in an event).
        
    \item \textbf{Architectural ablation experiments} - these are described in Section~\ref{sec:experiments}, for ease of readability.

\end{itemize}

%! Section 6 - Experiments and Discussion

\section{Experiments and Discussion}
\label{sec:experiments}

%----------------------------------------------------------------------

% It has been repeatedly observed that in some settings, random word embeddings perform as well as pretrained ones, 
% or very nearly, including in our baselines \cite{tilk2016event,hong-2018-learn-event-rep,marton-sayeed-lrec2022-RW-eng-v2}.
% We design experiments to answer the following questions: \textbf{Q1.} 
% Why is this so
% in our compositional semantics and psycholinguistic tasks? 
% \textbf{Q2.} For such semantic tasks and architecture, where is the learning encoded? Is it in the word embeddings, 
% role embeddings or “the network”?
% \textbf{Q3.} Training set size effect: is more data better for this indirect setting and tasks? 

    \begin{table*}[t!h!]
    \begin{threeparttable}
    \centering
    \begin{tabular}{|c|c|c|c|c|c|c|}
        \hline
        Embedding & Shared? & Tuned? & Role? & Role Accuracy & Word Accuracy & Epochs\tnote{*}\\
        \hline
        Random &         N  & Y & Y & $.9655 \pm .0014$ & $.1363 \pm .0020$ & $11(6)$\\ 
        Random & \textbf{Y} & Y & Y & $.9671 \pm .0003$ & $.1372 \pm .0022$ & $11(6)$\\ 
        GloVe  &         Y  & Y & Y & $.9669 \pm .0003$ & $.1374 \pm .0005$ & $15(10)$\\
        \hline
        \hline
        Random & Y & \textbf{N} & Y & $.6609 \pm .0046$ & $.1208 \pm .0012$ & $25(20)$\\ 
        GloVe  & Y &         N  & Y & $.9510 \pm .0011$ & $.1291 \pm .0006$ & $25(20)$\\ 
        \hline
        \hline
        GloVe & Y & Y & \textbf{NIR}\tnote{\dag} & $.9036 \pm .0013$ & $.1348 \pm .0019$ & $11(6)$\\
        GloVe & Y & Y & \textbf{NTR}\tnote{\dag} & $.9677 \pm .0006$ & $.1230 \pm .0017$ & $12(7)$\\
        GloVe & Y & Y & \textbf{NR}\tnote{\dag} & $.9007 \pm .0021$ & $.1078 \pm .0010$ &  $8(3)$\\
        \hline
        \hline
        RAND Network\tnote{\ddag}    & Y & Y & Y & $.1530 \pm .0716$ & $.0000 \pm .0000$ &   - \\
        Simpler Network\tnote{+} & Y & N & Y & $.9987 \pm .0005$ & $.1208 \pm .0020$ & $6(1)$\\
        \hline 
    \end{tabular}
    \caption{\label{table-1-1perc} 
        Word and Role accuracy on 1\% training data.} %(Epochs in parentheses: the epoch of the effective model (best model before early stopping after patience limit))} \textbf{N1}=No input role (in context); \textbf{N2}=No target role (in prediction); \textbf{N3}=No role; \textbf{RAND Net}=Network with no training that uses previously fine tuned word/role embeddings as input; \textbf{Simpler Net}=Simpler Feed forward Network with previously fine tuned word/role embeddings as input. Epochs in parentheses: the epoch of the effective model (best model before early stopping after patience limit)}
    \footnotesize
    \begin{tablenotes}
    \item[\dag] \textbf{NIR}=No input role (in context); \textbf{NTR}=No target role (in prediction); \textbf{NR}=No role
    \item[\ddag] Network with no training that uses previously fine tuned word/role embeddings as input
    \item[+] Simpler Feed forward Network with previously fine tuned word/role embeddings as input
    \item[*] Epochs in parentheses: the epoch of the effective model (best model before early stopping after patience limit)
    \end{tablenotes}
    \end{threeparttable}
    \end{table*}
    \begin{table*}[t!h!]
    \begin{adjustbox}{width={\textwidth},totalheight={\textheight},keepaspectratio}
    \begin{tabular}{|c|c|c|c|c|c|c|c|c|c|}
        \hline
        Embed. & Shrd & Tuned & Role & Padó & McRae & GDS & Ferretti-Loc & Ferretti-Instr & Bicknell\\
        \hline
        Random &         N  & Y & Y & $.5474 \pm .0345$ & $.3231 \pm .0236$ & $.4485 \pm .0314$ & $.2611 \pm .0036$ & $.2282 \pm .0623$ & $.5260 \pm .1185$\\ 
        Random & \textbf{Y} & Y & Y & $.5280 \pm .0274$ & $.3384 \pm .0174$ & $.4388 \pm .0206$ & $.2532 \pm .1421$ & $.2266 \pm .0391$ & $.5000 \pm .0673$\\ 
        GloVe  &         Y  & Y & Y & $.5316 \pm .0320$ & $.3280 \pm .0177$ & $.4534 \pm .0209$ & $.2851 \pm .0301$ & $.2895 \pm .0258$ & $.5438 \pm .0370$\\
        \hline
        \hline
        Random & Y & \textbf{N} & Y & $.4396 \pm .0344$ & $.2838 \pm .0109$ & $.2841 \pm .0246$ & $.1767 \pm .0273$ & $.2086 \pm .0322$ & $.4781 \pm .0450$\\ 
        GloVe  & Y &         N  & Y & $.4941 \pm .0247$ & $.3090 \pm .0254$ & $.4349 \pm .0229$ & $.3011 \pm .0301$ & $.3439 \pm .0421$ & $.5563 \pm .0490$\\
        \hline
        \hline
        GloVe & Y & Y & \textbf{NIR} & $.5079 \pm .0587$ & $.3205 \pm .0580$ & $.4217 \pm .0472$ & $.3054 \pm .0791$ & $.2543 \pm .0796$ & $.6042 \pm .0896$\\
        GloVe & Y & Y & \textbf{NTR} & $.2400 \pm .0294$ & $.0937 \pm .0258$ & $.3845 \pm .0083$ & $.3071 \pm .0017$ & $.2621 \pm .0531$ &  $.5469 \pm .0388$ \\ 
        GloVe & Y & Y & \textbf{NR} & $.2496 \pm .1088$ & $.1139 \pm .0150$ & $.3385 \pm .0363$ & $.2955 \pm .1243$ & $.2668 \pm .0375$ & $.5885 \pm .0448$\\ 
        \hline
        \hline
        RAND %Net    
        & Y & Y & Y & $-.0001 \pm .1090$ & $.0109 \pm .1604$ & $.0365 \pm .0784$ & $.0165 \pm .1048$ & $-.0346 \pm .0785$ & $.4531 \pm .1027$\\
        Simpler %Net 
        & Y & N & Y &  $.3271 \pm .0555$ & $.2175 \pm .0294$ & $.3356 \pm .0345$ & $.1055 \pm .0259$ &  $.0459 \pm .1239$ & $.5365 \pm .0593$\\
        \hline 
    \end{tabular}
    \end{adjustbox}
    \caption{\label{table-2-1perc} 
        Thematic Fit tests on 1\% training data (same models as in Table~\ref{table-1-1perc})}
    \end{table*}
    
\subsection{Shared Embedding Layer}
\label{sec:shared-emb}
    We modify the network to use a single embeddings set shared between
    the input words and target word, by using a single index-to-embedding mapping layer -- and similarly a shared embedding-mapping layer for the input roles and target role (Figure~\ref{fig:v5}). 
    This change results %in a reduction of learned parameters 
    in 2x the training speed (Section~\ref{sec:methodology}) without degradation in performance: role accuracy remains at 96.6-96.7\%, word accuracy at 13.6-13.7\%, Padó at 52-54\%, McRae at 32-33\%, and so on
    (see first two rows in Tables~\ref{table-1-1perc} and~\ref{table-2-1perc}).
    Therefore we use the faster shared architecture for the rest of the experiments.
    % Since RW-Eng v2 is very large,  
    We train all models (until Section~\ref{sec:larger-data}) on a uniformly sampled  
    1\% subset, which is large enough to get indicative results while saving time and cost in experimentation.  
    For comparison of our results to previous work, see Section~\ref{sec:larger-data}.

    \subsection{Random vs. Pre-trained Embeddings}
    \label{sec:pretrained-vs-random-emb}
    \citet{hong-2018-learn-event-rep} used random Glorot uniform to initialize the word embeddings. Private communication with the authors confirmed  random embeddings do as well as pretrained ones for these tasks.
    We replicate this finding, comparing random word embeddings to  
    pretrained GloVe embeddings \cite{pennington2014glove}, both of size 300: role accuracy at 96.7\%, word accuracy at 13.7\%, Padó at 52.8-53.2\%, McRae at 32.8-33.8\%. Similar trend follows across all the thematic fit task results with  overlapping 95\% confidence intervals of the experiments with random and GloVe embeddings 
    (rows~2 and~3 in the top part of Tables~\ref{table-1-1perc} and~\ref{table-2-1perc}).

    (\textbf{Q1}) Why is this so? 
    We note that during training,  embeddings get updated.
    To check if this update is responsible for bridging the gap between zero knowledge (random embeddings) 
    and much knowledge (compressed in the pre-trained GloVe),
    we freeze the word embedding layer and rerun the 
    experiments (see the middle part in the same two tables). 
    Contrary to our previous experiment, we find fixed GloVe embeddings do much better than 
    fixed random embeddings on all our tasks. 
    We also see tuning helps the model converge much faster (from 25 epochs down to 11-15). 
    
    We conclude that indeed much of the learning is captured in the word embeddings. Tuning them even on only 1\% of our training data bridges the knowledge gap from the pre-trained embeddings almost completely 
    %\sh{along with a faster convergence almost by 10 epochs} 
    (with possible exceptions on Ferretti and Bicknell).
    But we note that although lower, the fixed embeddings results are not near-random. 
    This leads us to (\textbf{Q2}) Where else is learning done, and to what extent?

\subsection{Role Contribution}
    
    We now turn to role ablation tests.
    First we take away the input roles from the context embeddings and call this the no-input-roles network \textbf{NIR} (see Figure~\ref{fig:v6_n1} and  the third part of Tables~\ref{table-1-1perc} and~\ref{table-2-1perc}).
    We do not see large drops in word prediction (from 14.7\% to 13.5\%), or thematic fit tasks such as Padó (from 53.2\% to 50.2\%) and McRae (32.1\% to 32.8\%),  except role prediction (from 96.7\% to 90.4\%), which we expect by construction. 
    Note that when predicting the target word, the NIR network still receives the target role information, which, together with at least the predicate, is likely often sufficient information for prediction.
    
    We find it surprising that input role ablation barely affects performance on the
    psycholinguistic tasks.
    Why is that? 
    One possibility:
    %is that 
    the input role contribution is negligible. 
    But another possibility is that in NIR, all (or almost all) the role 
    information was \textit{crammed} into the target role embeddings.
    To tease these apart, 
    we next take away 
    the target role from the penultimate layer of the network, but leave 
    the input roles intact. We call this no-target-role network \textbf{NTR} (see Figure~\ref{fig:v6_n2} and the row after NIR in the same tables).
    Now the role accuracy goes back to the base level of 96.7\% (as expected by construction),
    but word accuracy drops (from 13.7\% to 12.3\%) and so does performance on the psycholinguistic tasks, e.g., Padó (from 53.2\% to 24\%), McRae (32.8\% to 9.4\%). 
    We conclude that target role carries more crucial information than input roles for our psycholinguistic tasks,
    and that role information \textit{cramming}, 
    if it happens in NIR, does not happen in the other direction (NTR). 
    
    Finally, for completeness, we remove all role information from 
    the network.  
    We call 
    this no-role network \textbf{NR} (see Figure~\ref{fig:v6_n3} and same tables).
    This results in a drastic drop in word accuracy (from 13.7\% to 10.8\%) in addition to degradation of role accuracy from \textbf{NIR} experiment as well as the psycholinguistic 
    tasks (Padó falls from 53.2\% to 25\%, McRae from 32.8\% to 11.4\%,  and so on). 
    This is an 
    an interesting finding which supports previous knowledge about the importance of roles in multi-task learning setting while at the same time defies the importance of roles in the context vector (the output of the residual block in Figure~\ref{fig:allnets}). Next, we turn to learn more about the impact this vector and the block it is in.

\subsection{\enquote{It's the Network!}... Or is it?}
 
    In order to see how much the particular \ac{ResRoFA-MT} model architecture (aka \enquote{the network}) contributes in our tasks,  we first
    use the fine-tuned GloVe embedding from a previously trained base model 
    (third row in Table~\ref{table-1-1perc}) and assign the rest of the network  
    random weights (\enquote{RAND Network} in Tables~\ref{table-1-1perc} and~\ref{table-2-1perc}).
    To ensure the random weights are similar in size to the trained weights, we calculate the mean and
    standard deviation for each layer separately and assign that layer random weights  using a 
    Gaussian distribution with the same parameters.
    We see this new model does very poorly,  near random prediction (word accuracy at 0\%, role accuracy at 15.3\%, Padó at 0\%, McRae 1.1\% and so on). 
    This could be due to the learned representation in the network weights that were 
    ablated here but also due to incompatibility of the non-trained random network weights with the very informative word embeddings.
    
    Therefore next we replace the complex middle \textit{residual block} 
    with a plain 
    dense 
    %projection 
    layer but let this \enquote{Simpler Network} (Figure~\ref{fig:v9}, Tables~\ref{table-1-1perc} and~\ref{table-2-1perc}) learn during training. 
    In training here we use the fine-tuned word (and role) embeddings from our base model. 
    Curiously, we see a notable jump in role accuracy (from  96.7\% to 99.9\%), but a drop in 
    word accuracy (from 13.7\% to 12.1\%) as well as in the psycholinguistic tasks (Padó goes down from  53.2\% to 32.7\% , McRae from 32.8\% to 21.8\%, etc.) other than Bicknell's (53.7-54.4\%).  
    We speculate the latter task is an outlier here because it involves comparing the plausibility of two two-participant events with one participant changed.  A simpler network may have an easier time representing binary distinctions within a pair of simple events, as opposed to predicting fine-grained scores of more complex inter-relationships, evaluated with Spearman's $\rho$ in the other datasets. 
    It may even be able to rely on general collocation statistics here, regardless of roles, but we leave this for future work.
    Note that here, we still do multi-task prediction as before, but in a much simpler network. 
    
    This, along with the role ablation experiments,
    suggest that while the potential incompatibility of the non-trained random network weights 
    with the word embeddings may account for some of the drop in performance, 
    the context vector 
    formation through multiplication and likely also the improvements implemented in our 
    base model have a large impact on the representation learning as tested on 
    the thematic fit tasks (although not the same impact on word/role prediction). 

    We see again that there is no clear correlation between the increase in directly optimized 
    for word/role prediction, and the performance on the psycholinguistic tasks for which the models were not directly optimized.

    To recap, it seems the answer to (\textbf{Q2}) is nuanced:
    Padó and McRae are most sensitive to ablated roles;
    GDS, and perhaps  Bicknell, to non-tuned random word embeddings; 
    Ferretti to ablated (simplifiled) networks;
    and all are sensitive to RAND Networks, but Bicknell is surprisingly robust even there.

\subsection{Training Data Size Effect}\label{sec:larger-data}
    Often in machine learning and \ac{NLP}, models learn better with more data. 
    However, there are typically diminishing returns. 
    To test the effect of training data size, 
    we use our shared layer network with 
    tuned GloVe embeddings (as in row 3 in Table~\ref{table-1-1perc}) 
    on uniformly sampled 1\%, 10\%, 20\% 40\% and 100\%  of the training dataset. See Table~\ref{table-3-large} and Table~\ref{table-4-large}.

    \begin{table} [bth]
    \begin{threeparttable}
    \centering
    \begin{small}
    \begin{tabular}{|c|c|c|c|}
        \hline
        Sys   & Role Accuracy     & Word Accuracy     & Epochs\\
        \hline
        B1\tnote{\dag}    & \multicolumn{1}{l|}{$.9470$}              & -                 & -     \\
        B2\tnote{\ddag}    & $.9715 \pm .0010$ & $.1541 \pm .0045$ & -     \\
        \hline
        20\%M\tnote{+} & $.9707 \pm .0002$ & $.1450 \pm .0004$ & -     \\
        \hline\hline
        0.1\% & $.9446 \pm .0015$ & $.0994 \pm .0024$ & $12(7)$ \\
        1\%   & $.9669 \pm .0003$ & $.1374 \pm .0005$ & $15(10)$\\
        10\%  & $.9701 \pm .0002$ & $.1443 \pm .0006$ & $13(10)$\\ 
        20\%  & $.9703 \pm .0004$ & $.1445 \pm .0009$ & $9(6)$ \\
        40\%  & $.9704 \pm .0007$ & $.1442 \pm .0011$ & $9(6)$  \\
        100\%\footnote{100\% experiments are comprised of 2 trials while the rest are comprised of 5.} & $.9708 \pm .0006$ & $.1444 \pm .0019$ & $7(4)$\\
        \hline
    \end{tabular}
    \end{small}
    \caption{\label{table-3-large} 
        Comparison of performance 
        with GloVe (tuned) with varying training set sizes (Sys) 
    }
    \footnotesize
    \begin{tablenotes}
    \item[\dag] \citet{hong-2018-learn-event-rep} 20\%
    \item[\ddag] \citet{marton-sayeed-lrec2022-RW-eng-v2} 20\%
    \item[+] The average of max value in each trial for fair comparison  with benchmarks B1,B2
    \end{tablenotes}
    \end{threeparttable}
    \end{table}
    %\FloatBarrier
    
    \begin{table*}[th]
    \begin{threeparttable}
    \begin{adjustbox}{width={\textwidth},totalheight={\textheight},keepaspectratio}
    \begin{tabular}{|c|c|c|c|c|c|c|}
        \hline
        System & Padó              & McRae             & GDS         & Ferretti-Loc     & Ferretti-Instr   & Bicknell          \\
        \hline
        B1     & \multicolumn{1}{l|}{$.5300$}             & \multicolumn{1}{l|}{$.4250$}            & \multicolumn{1}{l|}{$.6080$ }            & \multicolumn{1}{l|}{$.4630$ }            & \multicolumn{1}{l|}{$.4770$ }            & \multicolumn{1}{l|}{$.7450$ }            \\
        B2     & $.5363 \pm .0035$ & $.4322 \pm .0232$ & -                 & -                 & -                 & -                 \\
        \hline
        20\%M  & ${.5855} \pm .0101$ & $.4338 \pm .0181$ & $.5495 \pm .0220$ & $.3539 \pm .0239$ & $.4255 \pm .0210$ & $.6094 \pm .0000$ \\
        \hline\hline
        0.1\% & $.2992 \pm .0441$ & $.1856 \pm .0157$ & $.1699 \pm .0180$ & $.0891 \pm .0306$ & $.0367 \pm .0203$ & $.4906 \pm .0402$ \\
        1\%    & $.5316 \pm .0320$ & $.3280 \pm .0177$ & $.4534 \pm .0209$ & $.2851 \pm .0301$ & $.2895 \pm .0258$ & $.5438 \pm .0370$\\
        10\%   & $.5572 \pm .0247$ & $.3993 \pm .0137$ & $.5409 \pm .0150$ & $.3410 \pm .0358$ & $.3765 \pm .0320$ & $.5906 \pm .0320$\\ 
        20\%   & $.5241 \pm .0558$ & $.3708 \pm .1182$\tnote{\dag} & $.5245 \pm .0148$ & $.3191 \pm .0312$ & $.3853 \pm .0454$ & $.5813 \pm .0210$ \\
        40\%   & $.3662 \pm .1355$ & $.3831 \pm .0276$ & $.5467 \pm .0183$ & $.3331 \pm .0215$ & $.3660 \pm .0284$ & $.5750 \pm .0460$ \\
        100\%\tnote{\ddag}  & $.3375 \pm .7293$ & $.3733 \pm .5203$ & $.5338 \pm .1328$ & $.2736 \pm .7846$ & $.3416 \pm .3297$ & $.6094 \pm .1985$ \\
        \hline
    \end{tabular}
    \end{adjustbox}
    \caption{\label{table-4-large} 
        Thematic Fit 
        with GloVe tuned (same models as in Table~\ref{table-3-large})
    }
    \footnotesize
    \begin{tablenotes}
    \item[\dag] 1 trial had an outlier score .2026
    \item[\ddag] All experiments had 5 runs per training subset, except for the 100\% with only 2 runs, due to compute resource limitation.
    \end{tablenotes}
    
    \end{threeparttable}
    \end{table*}

   First, in order to compare fairly with previous work, we report the average of the \textit{maximum} value in 
    each training trial on 20\% of the data. 
    (Recall that our 20\% of the data is a larger training set than our baselines' 20\% due to improvements in our batcher).
    Our role accuracy (97.1\%) is better than \citet{hong-2018-learn-event-rep} (94.7\%) and similar to \citet{marton-sayeed-lrec2022-RW-eng-v2} (97.2\%). 
    Our word accuracy (14.5\%) is a bit lower than the latter (15.4\%). 
    On the indirectly supervised thematic fit tasks, our results are better on Padó (58.6\% compared to 53\%), similar on McRae (42.5-43.4\%), 
    but lower for the rest.
    We suspect that in the previous work authors reported the \textit{best} of all the epochs
    from all trials, which can explain why the previously reported scores are higher than our results; but we could not verify that.

    In order to better understand the effect of training set size (\textbf{Q3}), we use next what we believe to be more
    realistic numbers: the average of the last saved model in each run (best model per our validation set) in each training subset size.  
    
    We see incremental improvements from the 0.1\% subset (role accuracy at 94.5\%, word accuracy at 9.9\%) to the 1\% subset (role accuracy at 96.7\%, word accuracy at 13.7\%) to the 10\% subset (role accuracy at 97.0\%, word accuracy at 14.4\%)   across all our evaluation tasks; 
    however, contrary to our null hypothesis, we see diminishing returns or no gains in role and word prediction when using 
    20\% 
    %(word accuracy at 97\%, role accuracy at 14.5\%) 
    or more 
    of the training set. 
    In most of the psycholinguistic tasks (Table~\ref{table-4-large}), results plateau at 10\% or 20\% (GDS at 52.5-54.1\%, Ferretti-Loc at 32-34.1\%, Ferretti-Instr at 36.6-38.5\% and Bicknell at 57.5-59.1\%) with the notable exception of Padó (best at 55.7\% with 10\% training data) and McRae (best at 40\% at 10\% training data) , where we see a negative trend at and beyond 20\%. 
    Why is it so, and only for these two tasks, with mainly Padó?
    \commentout{
    Padó: \ac{WSJ} high freq verbs, in them hi freq subj/obj;
Meant for NLP, no attempt to norm psycho-linguistically.
McRae: a bit more balanced than Padó, drawn from a freq distrib of a largret corpus, and not intended for \ac{NLP}, unlike Padó.
The rest of the psycholinguitic tasks solely depends on target word prediction.
    }
    The Padó dataset is constructed from high-frequency fillers. It behaves differently from the other datasets and gets a high maximum average score on the 20\% subset probably because there is more training data available for high-frequency fillers, compared to the other datasets, including McRae.
    Considering the small samples in these test sets, they might quickly become victims of not only high variance, but also of overfitting, that is to say, the models may specialize on the corpus distribution, increasingly with training set size. This distribution is likely to be different from the WSJ distribution, from which Padó dataset is drawn
    (but see also Section~\ref{sec:localcorrelation}). 

    How do word/role prediction and thematic fit tasks relate to each other? 
    We leave this question for  future research, but our hypothesis is that psycholinguistic 
    meaning of natural language is grounded in interaction with other modalities (e.g., actions, vision, audio), which a model cannot learn just from more textual training data. 
    
    This leads potentially to a much bigger question: how much can a neural model  learn natural 
    language by just being trained on very large corpora or billions of parameters, and where is the saturation point? 
    Furthermore, we see role information is important to our psycholinguistic tasks; how much does the role definition and granularity (e.g., PropBank or FrameNet), or the role set size, matter for these tasks?
    Possibly, with a richer roleset, we may see more alignment between word/role prediction and the psycholinguistic tasks. 
    Perhaps PropBank roles are too coarse-grained to allow for an analysis of how a role-prediction task relates to a thematic fit task, which involves the fine-grained ranking (via Spearman's $\rho$) of event plausiblities derived from the underlying semantic characteristics of the nouns and verbs involved. If so, understanding how performance on a role-prediction task relates to thematic fit judgements may not be possible without a finer-grained inventory of semantic characteristics, such as Dowtyan proto-roles \citep{dowty1991thematic}.

\subsection{Global and Local Correlation}\label{sec:localcorrelation}
We evaluate both Padó and McRae by computing Spearman's rank correlation between the sorted list of model's probability scores and the sorted list of averaged human scores, for each dataset.
Why do Padó and McRae  deteriorate with increasing training data size? To test if this is due to fluctuation of model scores for unrelated but near-in-score verb-noun pairs, we averaged correlations for \textit{local} subsets, grouped by verb. This should be an easier task, since some of the globally close competition is not present in each by-verb subset.
Indeed, we see high jumps of 5-8\% for
the \textit{local} correlation scores in the larger subsets (40\% and 100\%). But in the smaller subsets we see changes of 2-3\% up or down. Moreover, 
 the trend of lower correlation with larger training sets remained.
We leave it to future work to dig further into why Padó and McRae show such an anomaly.

%! - Section 6 Conclusion

\section{Conclusions and Future Work}
\label{sec:conclusion}

%----------------------------------------------------------------------

In this work, we explored 
 why random word embeddings counter-intuitively perform as well as pretrained 
word embeddings on certain compositional semantic tasks (some being outside the  models' explicit objective),  
where the learning is actually stored (teasing apart the word 
embeddings, role embeddings, and the rest of the network),
and how training set size affects performance on these tasks.
We found out that tuning (or further tuning) the word embeddings helps and can bridge the gap between random and pretrained embeddings. Moreover,
our tuned embedding space  is different 
from pretrained embeddings like GloVe. 
We saw that the target role is more important 
than the input roles on our tasks. Furthermore, our experiments suggested that much of the learning happens also in the rest of the network outside word and role embedding layers. 
No single factor (word and role embeddings or the network) is most important for all tasks.

Training set size had a surprising negative effect on Padó and McRae beyond 20\% of the training data. We attempted explaining this with an alternative evaluation method, but this remains to be explained further.

We release our code, including our preferred network architecture -- a modified version of  \ac{ResRoFA-MT}  with shared embedding layers.

One avenue in which we want to invest is to better understand the complex relationship between word/role accuracy and our psycholinguistic tasks. 
While our initial hypothesis was that training the network to minimize loss 
 on word/role prediction would also optimize performance on all our  tasks, 
 this did not always hold. 
We suspect that the groundedness is the missing link for (artificially and naturally) learning psycholinguistic tasks, and therefore adding grounding seems promising to us.

Another future avenue is to investigate the high variability in   psycholinguistic task performance compared to the fairly stable results on the directly optimized-for word and role prediction tasks.
%! - Section 8 Limitations (required by EMNLP 2022)

\section*{Limitations}
\label{sec:limitations}

There are certain limitations that were unavoidable in this work. One of them is the limited size of the available training and evaluation datasets for testing thematic fit tasks. It is likely that the high variance we observed is due to both our indirect supervision approach (in part due to lack of directly relevant data for training), and the small-size test sets. 
We are limited here by the state of the art in such datasets, not just by their size. It is a complex task to create and evaluate thematic fit with full phrases and sentences, i.e., not just with the arguments' syntactic heads. Since we do not know of any such datasets, our model was designed with only syntactic heads in mind.

Another limitation is the training dataset quality: due to its size, the training data was machine-annotated (for syntactic parsing, SRL and lemmas) and therefore unintended noise and bias may have been introduced in the models. In addition, even though our training datasets were collected with the goal of making them domain-general and balanced, it is hard to enforce and verify that in large sizes. 
We take issues such as toxicity and gender bias seriously, but we think that in our settings, where the model does not generate language and the test sets do not involve gendered examples, the related risks approach zero.

Semantic tasks such as thematic fit would most likely benefit from training on grounded language, e.g., combining text and vision, but working with such datasets is beyond the scope of this work.

Finally, a rather trivial limitation we have is the number of trials per experiment we could run due to time and computational constraints. We only ran 3-5 trials per experiment but a larger number of trials may yield more robust results. Despite all these limitations, we believe our work gives a very comprehensive analysis of the ResRoFaMT model and opens up some interesting avenues for future research work.

%! - Section 7 Ethical Considerations

\section*{Ethical Considerations}
\label{sec:ethics}

%----------------------------------------------------------------------
Our work uses RW-Eng v2 \cite{marton-sayeed-lrec2022-RW-eng-v2}, which in turn uses two corpora: ukWaC and the \ac{BNC}. Therefore, we have similar ethical concerns as mentioned in that previous work, including the way the BNC data was collected. Those who so wish can easily exclude the BNC data (it comprises only a small part of the whole corpus) and retrain.

The RW-Eng corpus (v1 or v2) could introduce undesired bias in use outside the UK, since the data is sourced entirely from UK web pages and other UK sources from the 20th century. English used outside the UK, and more recent English anywhere,  differ from this corpus in their word distributions, and therefore their input may yield sub-optimal or undesired results. Furthermore, models trained on it could encode a Western-centric view of
the world. 

The silver labels -- the automatic parsing and tagging of the corpus -- could introduce bias from the parsing / tagging algorithms. These parsers / taggers are
also trained models, which could be affected by their data sources. If this is a concern for some users, we encourage them to perform  validation of the data and its annotations.

Having said that, we believe that for most if not all conceivable applications, especially as long as one keep these limitations in mind, our work should not pose any practical risk.
%! Author = mughi
%! Date = 2/25/2022

\section*{Acknowledgements}
    \anon{
        This work started as a Masters Capstone project
        at the Columbia University Data Science Institute.
        We would like to thank Data Science Institute, and
        the Department of Computer Science at Columbia
        University for their support. 
        Specifically, Smaranda Muresan for early discussions and  Eleni Drinea for
        timely support with our computing resources in the initial stages of this project.  
        We would like to thank
        Google Cloud Platform for an award of credits to
        Asad Sayeed, allowing us to use Google Cloud
        computing resources. We are grateful to the rest
        of the Capstone team members – Jake Stamell, Anjani Prasad Atluri and Priyadharshini Rajbabu – for
        their valuable contributions. 
        This research was funded in part by
        a Swedish Research Council (VR) grant (2014-39)
        for the Centre for Linguistic Theory and Studies in
        Probability (CLASP). 
    }

\bibliography{anthology,custom, mtrf, thematic-fit}
\bibliographystyle{acl_natbib}

%\break
\vspace{2cm}
\section*{Appendix A: Why Not  Use BERT Here?}
\label{sec:bert}

With the advent of contextual embeddings, static word embeddings are often regarded as inferior or outdated. While this is true in many cases, we wish to point out that 
\textbf{(a)}~there are still cases where static word embeddings outperform BERT (\citet{lenci2022comparativeDSM,henlein2022toothbrushes}, \textit{inter alia}) 
and more importantly, 
\textbf{(b)}~not all NLP tasks and test sets are in the form of complete sentences, which may render contextual models useless there.
More specifically for our tasks:

\begin{enumerate}
    \item  As we point out at the end of Section~\ref{sec:related-work}, 
    %in the paper, 
    ``\citet{lenci2022comparativeDSM} demonstrate that ... even recent contextual models such as BERT are not necessarily better for out-of-context tasks than well-tuned static representations, predict or otherwise." 
    Our tasks, as represented by several psycholinguistic test sets are such out-of-context tasks. This \textit{is} the state-of-the-art in psycholinguistics datasets. The human judgments in these sets were given without a full sentence or other context beyond the verb-noun (or noun-verb-noun) items. There is no reason to assume that BERT, a contextual model trained on full sentences will do well on out-of-context tasks, and again, it has been shown BERT is not necessarily better than static word embeddings.

    \item  Even if we wanted to use BERT, we cannot use a lookup table the same way we can for static word embeddings. This means we will either have to decode each sentence on the fly every training (and evaluation) iteration, or decode once and save on disk. Simple calculation shows that the required storage demands, even for, say, 1\% of the data, make this exercise computationally extremely expensive.

    \item  BERT may break words to several tokens. How to map these to the verb or noun in the training or test is not always straightforward, and this mapping makes embedding extraction speed 0.5x slower. 

    \item  ``Hallucinating'' synthetic sentences from the verb-noun input in order for BERT to receive a sentence for input would invalidate the ratings given by the human raters without these (or other) sentences.

    \item  In order to validate our claims here, we experimented with  BERT on-the-fly in preliminary studies, using a small training subset of a few thousand sentences with simple token mapping, and the results were dismal while the training already excruciatingly slow. 

    \item  There is nothing wrong with systematically exploring models that use static word embeddings, even if contextual embeddings excite many people more. We don't think we should defend this choice.

\end{enumerate}

\section*{Appendix B: Dataset Examples}
%The full training and test dataset is available on request. However, below we try 
For added clarity we give the readers a few examples of  the training and evaluation data.
\begin{enumerate}
    \item Each training  example is a list of \textit{word/role} pairs. Each of these examples are created with the head words from full sentences, i.e.,  our final training examples do not contain full context. We have 6 semantic roles such as \textit{Arg0} and \textit{Arg1} etc. + one placeholder \textit{UNK} role for roles not included in our role set. A typical example could look like this: \texttt{\{0: 6, 1: 97, 2: 43511, 3: 43511, 4: 239, 5: 143, 6: 64\}}, where the numbers are pairs of role:word indices. The size of word vocabulary of the 0.1\% training subset is 43,510 with tokenId 43,510 and 43,511 corresponding to \textit{UNK} and \textit{Missing} word (the corresponding role does not have a head word). Larger training subsets vocabulary is capped at 50,000.

    Each example is used in training 1 or more times, each time with a different target word/role pair (see Figures~\ref{fig:v4}, \ref{fig:v5}, and Section~\ref{sec:methodology}), while the rest of the pairs are used for input. Note that a pair with a \textit{Missing} word cannot serve as a target word/role.
    
    \item Apart from the train / test split, we use multiple psycholinguistic evaluation sets that we do not optimize the model on, as mentioned in Section~\ref{sec:datasets}. While they all vary, a typical example is
    \texttt{\{client, advise, Arg0, 3.7\}}, which means that human raters gave an average of 3.7 to 'client' as Argument 0 (typically Agent) for 'advise' (as in ``the client advised the banker that ...''). In contrast,
    \texttt{\{client, advise, Arg1, 6.6\}}, means that human raters gave an average of 6.6 to 'client' as Argument 1 (typically Theme/Patient) for 'advise' (as in ``the banker advised the client that ...''). According to these human raters, 'client' fits semantically much better as Arg1 than Arg0 for 'advise'. 
   During thematic fit evaluation, we sort these test examples by human rater average scores, and the model output by model score. Then, we compute Spearman's rank correlation between the two sorted lists, as explained in Section~\ref{sec:localcorrelation}.
\end{enumerate}

% \section*{Appendix C: Are We Over-fitting on the Training Data?}
% \begin{center}
%  \begin{table} [bth]
%     \begin{threeparttable}
%     \centering
%     \begin{small}
%     \begin{tabular}{|c|c|c|c|c|}
%         \hline
%         Data Size   & 10\% & 20\% & 40\% & 100\%\\
%         \hline
%         Val Loss\tnote{\dag} & $7.0086$ & $5.7149$ & $5.7062$ & $5.6997$ \\
%         \hline
%     \end{tabular}
%     \end{small}
%     \caption{\label{table-i} 
%         Comparison of validation loss 
%         across different training data size
%     }
%     \footnotesize
%     \begin{tablenotes}
%     \item[\dag] Minimum validation loss during training
%     \end{tablenotes}
%     \end{threeparttable}
%     \end{table}
%     \end{center}

% We investigate if with larger of training data size beyond 20\% led to  over-fitting that may explain the negative trend in Padó and McRae. However, we note here that, Padó and McRae are not the direct tasks on which the model training is optimized rather the training objective is to maximize the word/role accuracy on the validation dataset. From Table\ref{table-i} we see validation loss does not change significantly beyond 20\% dataset which explain the behaviour of word/role accuracy. This confirms the fact we are not over-fitting on training data. We try to answer this question from the perspective of underlying data in \ref{sec:localcorrelation}.

\end{document}